\documentclass[letterpaper, 10 pt, conference]{ieeeconf}  

\IEEEoverridecommandlockouts                              

\usepackage{graphicx}
\usepackage[hidelinks]{hyperref}
\usepackage{amsmath} 

\title{\LARGE \bf
Traffic Regulation-aware Path Planning with Regulation Databases and Vision-Language Models
}

\author{Xu Han$^{1}$ Zhiwen Wu$^{1}$ Xin Xia$^{1}$ and Jiaqi Ma$^{1}$
\thanks{$^{1}$Xu Han, Zhiwen Wu, Xin Xia and Jiaqi Ma is with the UCLA Mobility Lab, University of California Los Angeles, 4731G Boelter Hall, Los Angeles, CA, USA. Corresponding author: {\tt\small jiaqima@ucla.edu.}}%
}

\begin{document}

\maketitle
\thispagestyle{empty}
\pagestyle{empty}

\begin{abstract}

This paper introduces and tests a framework integrating traffic regulation compliance into automated driving systems (ADS). The framework enables ADS to follow traffic laws and make informed decisions based on the driving environment. Using RGB camera inputs and a vision-language model (VLM), the system generates descriptive text to support a regulation-aware decision-making process, ensuring legal and safe driving practices. This information is combined with a machine-readable ADS regulation database to guide future driving plans within legal constraints. Key features include: 1) a regulation database supporting ADS decision-making, 2) an automated process using sensor input for regulation-aware path planning, and 3) validation in both simulated and real-world environments. Particularly, the real-world vehicle tests not only assess the framework's performance but also evaluate the potential and challenges of VLMs to solve complex driving problems by integrating detection, reasoning, and planning. This work enhances the legality, safety, and public trust in ADS, representing a significant step forward in the field.

\end{abstract}

\section{INTRODUCTION}

As Automated Driving Systems (ADS) advance, fully automated mobility becomes more attainable. Research has significantly improved sensor-based perception, vehicle control, and decision-making \cite{HAN2022103952}. Ensuring autonomous vehicles (AVs) comply with laws is crucial for safety, efficiency, and public acceptance. However, a key gap remains: the lack of a machine-readable regulation database for ADS software and comprehensive ADS-specific laws across jurisdictions.

To address this gap, the US Federal Highway Administration (FHWA) launched a prototype data framework for traffic regulations in 2021 \cite{fhwa2021data}, bringing stakeholders together to develop voluntary specifications. These specifications help standardize operations for infrastructure owners, ADS developers, and technology providers. Recent research has explored integrating legal constraints into ADS decision-making and path planning. Building on this, our regulation-aware framework combines a comprehensive ADS traffic law database with a finite state machine (FSM) \cite{mcculloch1943logical} and cost functions to evaluate plans based on safety, comfort, and legality. A VLM further enhances adaptability by directly interpreting driving conditions, reducing reliance on specialized models like object detection or trajectory prediction, which can struggle in certain scenarios. 

This paper's contribution is two-fold: (1) proposing a practical path planning framework that feasibly integrates with VLMs for traffic regulation-aware planning (as compared to direct VLM-based driving, which has been proved as unreliable) and (2) performing simulation and real-world vehicle experiments to assess the framework's performance and understand the potential and challenges of VLM in this critical use case for future reference.

\section{RELATED WORKS}

Recent research highlights the need for ADS systems to comply with traffic laws, recognizing the critical role of regulations in ensuring the safe and effective operation of autonomous vehicles. Ilková et al. \cite{7976252} provide an overview of AV legal frameworks in Europe and the United States, emphasizing the importance of understanding how legal provisions apply to ADS and the need for harmonizing regulations across jurisdictions. Bakar et al. \cite{su14031456} similarly stress the necessity of unified global guidelines to enhance roadway safety, as inconsistent traffic laws pose challenges for AV developers. While these studies emphasize legal accountability in ADS operations, current regulatory efforts have yet to develop machine-readable databases to support ADS decision-making fully. LEE and Hess \cite{LEE202085} also highlight progress in addressing national regulations but note the complexities in adapting AV software to comply with varying legal requirements across regions.

Efforts to integrate legal constraints into ADS decision-making and path planning are gaining momentum. Zhang et al. \cite{3460082} introduced a framework using formal methods to detect violations of scenario-based driving rules, providing a rigorous approach for ADS to comply with traffic regulations. Cho et al. \cite{8967708} proposed a deep learning model that predicts vehicle paths while assessing regulatory compliance. De Vries et al. \cite{de2022regulations} incorporated traffic regulations into a cost function within a local model predictive contouring control (LMPCC) system for real-time motion planning. Despite these advances, many methods are limited by focusing on a narrow set of regulations and lacking jurisdictional awareness. Additionally, frameworks like AVChecker and LMPCC often depend on specialized detection models and act on pre-selected sets of traffic regulations, reducing their real-world applicability. It is necessary to utilize the comprehension and reasoning capabilities for such complex tasks. 

\section{METHODOLOGY}

\begin{figure}[!ht]
  \centering
  \includegraphics[width=0.45\textwidth]{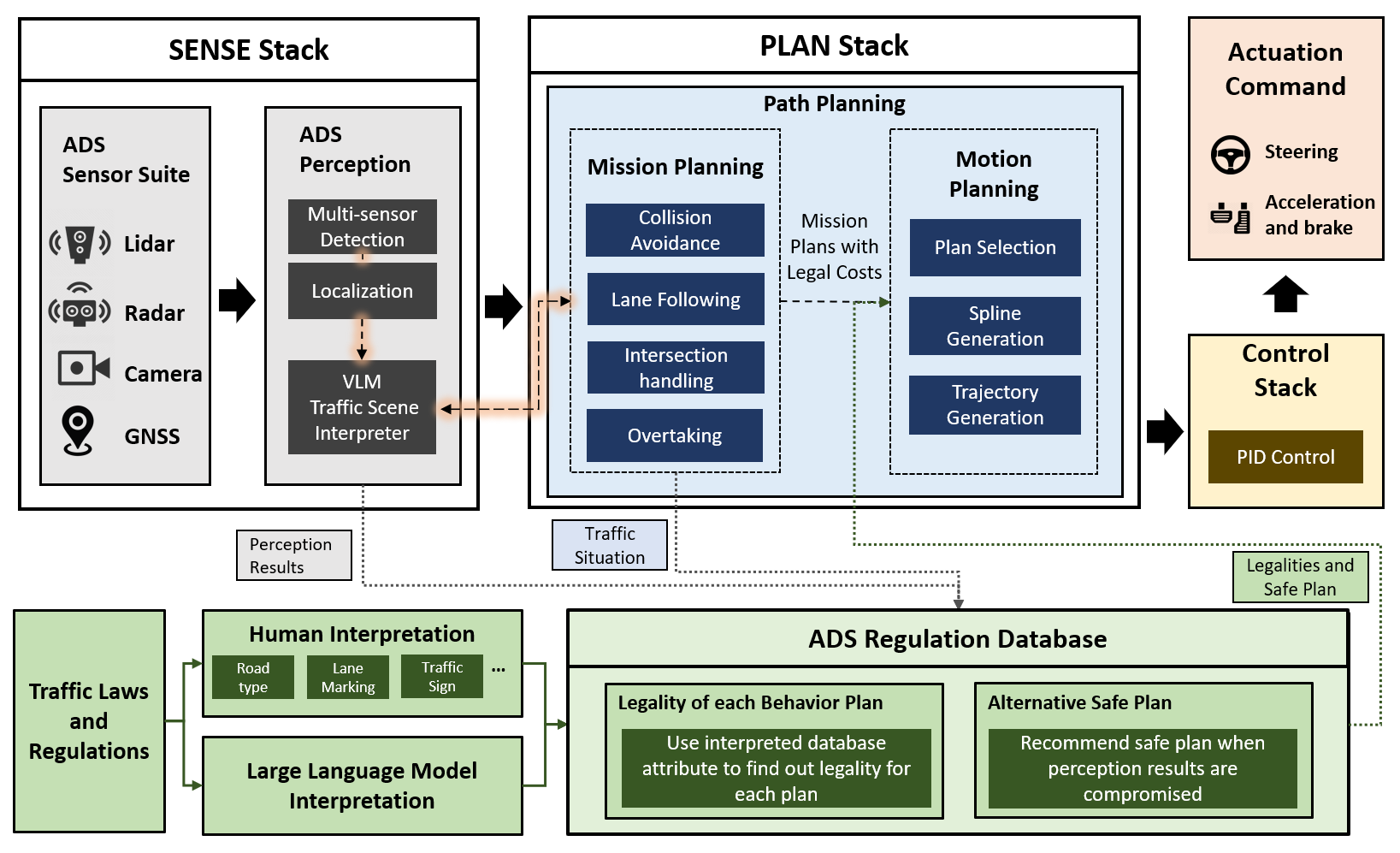}
  \caption{The overview of the path planning framework where the path planning modules are highlighted in blue, the interaction between VLM and mission planning is highlighted in orange, and the ADS regulation database is highlighted in green.}\label{fig:overview}
\end{figure}

\subsection{Framework Overview}
The proposed framework, shown in Fig. \ref{fig:overview}, integrates traffic laws into real-time ADS decision-making through mission (strategic) and motion (tactical) planning. In mission planning, the integrated FSM evaluates candidate driving plans based on real-time perception data. To ensure regulation compliance, the FSM leverages the VLM LLaVA \cite{liu2024llava} to interpret the driving scene and generate text-based responses. These responses are then compared against a structured regulation database to assess the legality of each candidate plan for the next step. This interaction, highlighted in orange in Fig. \ref{fig:overview}, enables more adaptive and regulation-aware decision-making.

After generating candidate plans in mission planning, motion planning creates a detailed trajectory with refined resolution using cubic spline interpolation \cite{mckinley1998cubic}. A cost function then selects the best option based on comfort, economy, safety, and legality, ensuring compliance with traffic laws while optimizing driving performance for real-time ADS planning.

\subsection{ADS Regulation Database}
Traffic regulations in the U.S. vary by state and locality, posing challenges for ADS. In the initial phase of this study, common elements from the Uniform Vehicle Code (UVC) \cite{national1952uniform}, California Vehicle Code (CVC) \cite{cvc_entire}, Los Angeles County Vehicle Code (LACVC) \cite{LACVC}, and City of Los Angeles Vehicle Code (CLAVC) \cite{CLAVC} were analyzed to create a foundational ADS database. This database transitions traffic legislation into a machine-readable format, accommodating diverse regulations across different jurisdictions.

To ensure regulation compliance, this study develops a structured, machine-readable ADS regulation database adaptable to local jurisdictions. Traffic codes are converted into a structured CSV format for efficient parsing and integration with ADS decision-making. Each entry includes metadata such as effective dates and legislative locations, along with possible current and next state fields aligned with the FSM in the proposed framework. This allows real-time vehicle states to be compared against regulatory requirements. Additionally, each code includes description features like “Condition” and “Result” for keyword matching and numerical features like “max speed” for direct comparison. As shown in Table~\ref{tab:spdlimiteg}, traffic codes are structured into attributes that help the perception module assess legality, transforming raw regulation text into actionable ADS parameters.

\begin{table*}[!ht]
	\caption{An example regulation of the traffic regulation database}\label{tab:spdlimiteg}
	\begin{center}
		\begin{tabular}{|p{3cm}|p{3cm}|p{1cm}|p{1cm}|p{1cm}|p{1cm}|p{1.4cm}|p{1.4cm}|}
            \hline
            \textbf{Code Text} & \textbf{Condition} & \textbf{Result} & \textbf{Legality} & \textbf{Attribute: Road Type} & \textbf{Attribute: Max Speed} & {Possible Current States} & {Possible Next States}\\
            \hline
            ``A person who drives a vehicle upon a highway at a speed greater than 100 miles per hour is guilty…'' & ``Vehicle driven upon a highway at a speed greater than 100 miles per hour.'' & ``is guilty...'' & FALSE & Highway & 100 mph & Car Following; Go Straight; Overtaking & Car Following; Go Straight; Overtaking  \\
            \hline
            \end{tabular}
	\end{center}
\end{table*}

\subsection{FSM Integration}
\begin{figure}[!ht]
  \centering
  \includegraphics[width=0.4\textwidth]{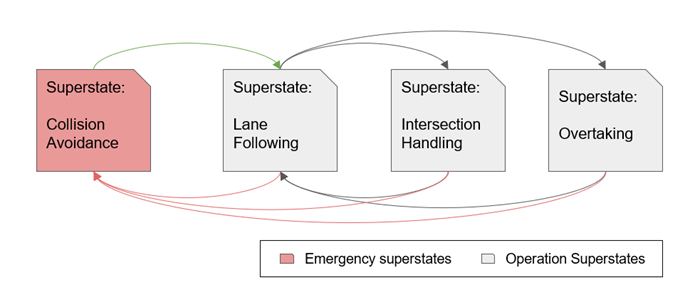}
  \caption{FSM superstates and transitions overview. Each superstate contains additional states and corresponding transitions.}\label{fig:fsm_ss}
\end{figure}

The decision-making framework of the proposed system is driven by a rule-based FSM designed to regulate vehicular behavior in complex driving scenarios. The FSM is organized into four super-states (as shown in Fig.\ref{fig:fsm_ss}): Lane Following, Intersection Handling, Overtaking, and Emergency Stop. Each super-state comprises multiple states and predefined transitions that encapsulate distinct driving behaviors. For example, the ego vehicle can start in the Lane Following mode, and if the navigation destination requires a turn, the FSM can switch to the Intersection Handling super-state provided that the vehicle was previously in the go-straight state within Lane Following. Once in Intersection Handling, the vehicle completes the turn by following transitions—such as switching between car-following and turn-right states—and then reverts to the Lane Following super-state.

The detailed path planning leverages the vector map incorporated by OpenCDA, which is accessible to all CDA applications within this ecosystem. Based on the current and intended states, the planning module selects target waypoints from the centerline coordinates of the corresponding road segments. A smooth path is then generated using cubic spline interpolation, ensuring that the trajectory conforms to the road structure while satisfying dynamic constraints, such as speed limits and safety metrics, thereby supporting safe and efficient navigation in dynamic traffic environments.

\subsection{VLM Integration}
\begin{figure}[!ht]
  \centering
  \includegraphics[width=0.5\textwidth]{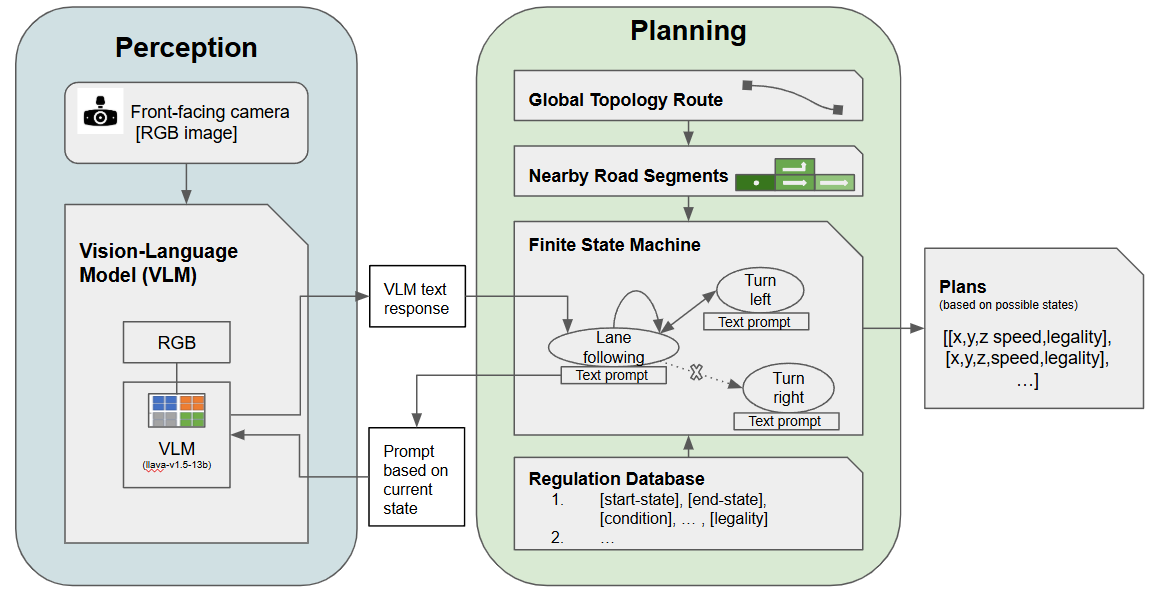}
  \caption{Framework structure of the VLM integrated path planning framework.}\label{fig:vlm_int}
\end{figure}

Planning is essential in ADS architecture for legal compliance and safety. This work integrates a VLM to generate high-level traffic descriptions, including intersections and signals, enabling a more coherent approach to regulation-aware planning. As shown in Fig.\ref{fig:vlm_int}, the VLM is embedded in the ADS perception pipeline, analyzing camera frames to assess the driving environment and generate text summaries of regulation-related conditions, which are then sent to the FSM. This process, called ``instruction tuning," leverages models like LLaVA—a multimodal system combining a vision encoder with a GPT-4-based Large Language Model (LLM) for enhanced visual and language understanding \cite{liu2024llava, ChatGPT}. 

The planning workflow begins with the global route, represented in a vector map format, which defines the sequence of road segments connecting the current location to the destination. As illustrated in Fig.\ref{fig:vlm_int}, the next possible road segments are extracted based on the current position, and the FSM selects the corresponding candidate states for the next maneuver. Simultaneously, the VLM processes visual input leveraging the front-facing camera feed and a context-specific prompt to generate a textual interpretation of the driving environment. All potential strategic plans for the next step are then aggregated and evaluated against the regulation database. The legality cost is computed by matching the current state, next state, and VLM-generated conditions with relevant regulatory constraints. This structured approach ensures that planned actions comply with traffic regulations while maintaining safe and efficient decision-making. 

Furthermore, to enhance decision-making, each FSM state is designed with a custom prompt to guide the VLM’s interpretation of the driving environment. As the foundation of strategic planning, the FSM defines states and corresponding actions while generating tailored prompts that focus the VLM on critical scene elements. For example, within the ``Lane-Following" superstate, prompts differentiate between sub-states such as ``Intersection Handling" and ``Overtaking", instructing LLaVA to ``examine the current driving scenario" or ``look out for intersections or obstacle vehicles". Table~\ref{tab:fsm_and_prompt} outlines the prompts associated with each superstate and their respective transitions. Additionally, the ``Emergency Handling" superstate, which is triggered in safety-critical situations, bypasses LLaVA and relies exclusively on distance measurements to ensure a timely and reliable response under critical hazard conditions.

\begin{table*}[!ht]
	\caption{Superstate Corresponding Text Prompts and Transitions }\label{tab:fsm_and_prompt}
	\begin{center}
		\begin{tabular}{|p{5cm}|p{5cm}|p{5cm}|}
            \hline
            \textbf{SuperStates} & \textbf{LLaVA Text Prompt
} & \textbf{Possible Next Superstates}\\
            \hline
            Lane Following & ``Examine the current driving scenario, look out for intersections or obstacle vehicles.'' & Lane Following; Intersection Handling; Overtaking \\
            \hline
            Intersection Handling & ``Examine the current driving scenario, check if the ego vehicle is still facing an intersection.'' & Intersection Handling; Lane Following \\
            \hline
            Overtaking & ``Examine the current driving scenario, check nearby lane occupation conditions, and look out for intersection.'' & Overtaking; Lane Following; Intersection Handling \\
            \hline
            \end{tabular}
	\end{center}
\end{table*}

\subsection{Cost Function}
The cost function evaluates travel plans by computing a total score based on four key components: legality, safety, comfort, and distance to the navigation goal, each weighted for behavioral tuning. The \textbf{legality cost} ($C_{\text{legal}}(P_i)$) is derived from a regulation database containing the vehicle's current FSM state, potential next states, conditions, and legality results. Since legality is binary, legal plans have zero cost, while illegal ones incur a penalty. The \textbf{safety} and \textbf{comfort costs} ($C_{\text{safety}}(P_i)$ and $C_{\text{comfort}}(P_i)$) relate to driving dynamics, considering average planned acceleration (including deceleration), speed variance, and maximum planned curvature. The total cost is the sum of these motion parameters, rewarding smoother plans with minimal fluctuations, steady speeds, and gentle turns. The \textbf{distance cost} ($C_{\text{distance}}(P_i)$) is computed in Frenet coordinates \cite{crenshaw1993orientation}, which measures the vehicle's advancement along the reference path to prioritize efficient routing toward the goal. The detailed calculations are presented below as (1) to (2): 

$$
C_{\text{total}}(P_i) = \mathbf{W} \cdot 
\begin{bmatrix} 
C_{\text{legal}}(P_i) \\ 
C_{\text{safety}}(P_i) \\ 
C_{\text{comfort}}(P_i) \\ 
C_{\text{distance}}(P_i) 
\end{bmatrix}
\eqno{(1)}
$$
$$
\mathbf{W} = \begin{bmatrix} w_{\text{legal}}, & w_{\text{safety}}, & w_{\text{comfort}}, & w_{\text{distance}} \end{bmatrix}
\eqno{(2)}
$$
where $C$ denote cost, and $W$ denote weight. During operation, each cost component is dynamically calculated using real-time vehicle data, enabling a data-driven evaluation of every plan. The final cost score guides trajectory selection, prioritizing the plan with the lowest weighted cost while ensuring legal compliance.

\section{EXPERIMENTS}
The experiments consist of two phases: simulation and real-world testing. The simulation phase validates the framework’s functionality using LLaVA as a perception tool to ensure regulation compliance in ADS vehicles, assessing the full ADS cycle with a focus on legality and safety. Real-world tests evaluate system performance in actual driving conditions, where LLaVA’s inference efficiency is critical for real-time decision-making.

\subsection{Simulation Test}
The simulation tests use UCLA’s OpenCDA platform \cite{xu2021opencda}, an open-source, full-stack ADS framework integrating perception, localization, planning, control, and V2X communication. The scenario simulates real-world complexity by incorporating multiple traffic regulations simultaneously. The ego vehicle begins in the rightmost lane of an intersection with a 35 mph speed limit, encounters a cyclist, makes a right turn, and enters a school zone with a reduced 25 mph limit, requiring it to navigate various challenges while ensuring compliance. In this simulation, three traffic regulations co-exist:
\begin{itemize}
    \item \textbf{Cyclist avoidance: }A driver of a motor vehicle shall not overtake or pass a bicycle proceeding in the same direction on a highway at a distance of less than three feet between any part of the motor vehicle and any part of the bicycle or its operator. California Vehicle Code Section 21760 \cite{cvc_21760}.
    \item \textbf{Right turn on red: }Except when a sign is in place prohibiting a turn, a driver, after stopping as required by subdivision (a), facing a steady circular red signal, may turn right. California Vehicle Code Section 21453 \cite{cvc_21453}.
    \item \textbf{School Zone: }A 25 miles per hour prima facie limit in a residence district, on a highway with a posted speed limit of 30 miles per hour or slower, when approaching, at a distance of 500 to 1,000 feet from, a school building. California Vehicle Code Section 22358 \cite{cvc_22358}.  
\end{itemize}

\subsection{Real-World Test}
\begin{figure*}[!ht]
  \centering
  \includegraphics[width=0.7\textwidth]{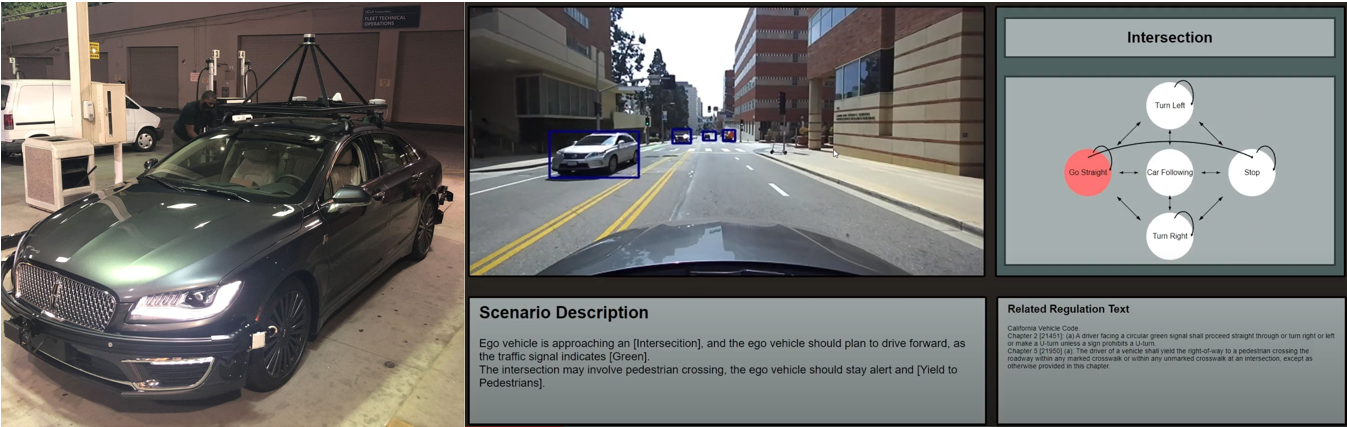}
  \caption{ADS vehicle and the real-time WebUI interface for VLM and FSM visualization.}\label{fig:v_and_UI}
\end{figure*}

Real-world testing assessed the regulatory data framework’s adaptability and compliance in dynamic traffic conditions. The test scenario was conducted at a UCLA campus intersection, involving the ego vehicle traversing the intersection and making a legal right turn on red when permitted. The ADS system, operating with full perception capability and a strategic planning module, identified key traffic elements, including signs, vehicles, and vulnerable road users (VRU) linked to regulations. The VLM’s ability to recognize lanes and speed limits was tested to support legal driving decisions. Due to regulatory and safety constraints, a human driver controlled the vehicle, eliminating the need for motion planning and detailed interpolated trajectories. The ADS system processed inputs from a front-facing camera and GNSS unit, with the VLM providing real-time text-based inferences to assist planning. A WebUI displayed FSM status, detection outputs, and VLM responses, while static planning was not visualized due to manual driving. The ADS vehicle and WebUI interface are shown in Figure~\ref{fig:v_and_UI}.
\section{RESULTS}
This section presents simulation and real-world test results for ADS regulation compliance. Efficiency is also evaluated to ensure safe, legal operation in real-world driving.

\subsection{Simulation Testing}
The simulation results demonstrate the robustness of the proposed path planning framework. In a complex scenario with three traffic regulations, the ego vehicle avoids a cyclist, stops at a red light before turning, and adjusts speed in a school zone. The VLM's inference and the framework's path planning were also tested against simulated human-driven traffic, demonstrating effective performance. Results are presented as downtrack distance over time, where downtrack distance, or station in the Frenet coordinate system \cite{crenshaw1993orientation}, represents the vehicle’s longitudinal position along the reference path. Its time derivative corresponds to longitudinal speed, offering insight into the vehicle’s behavior in each scenario.

\subsubsection{Overtaking Cyclist}
\begin{figure}[!ht]
  \centering
  \includegraphics[width=0.5\textwidth]{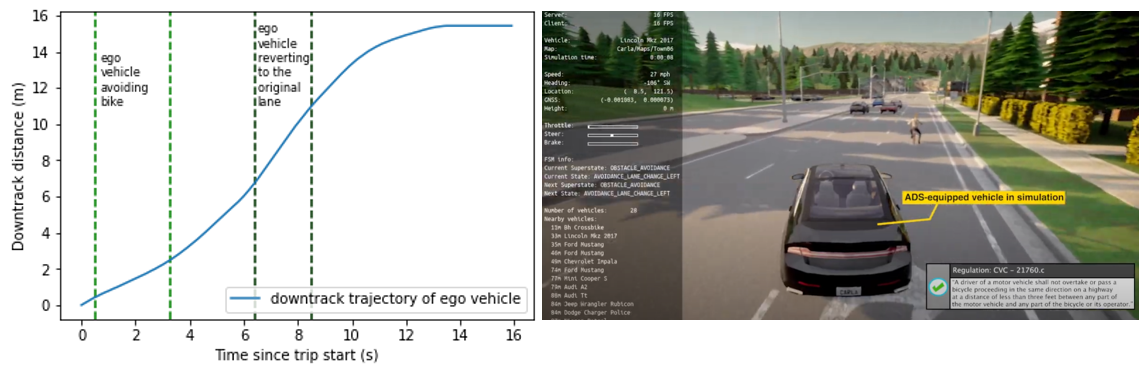}
  \caption{Trajectory plot and the simulation screenshot when ego ADS vehicle overtaking a cyclist.}\label{fig:avoid_bike}
\end{figure}

Two key events occur in this scenario: the ego vehicle detects the cyclist ahead and initiates overtaking, marked by the first moment, and completes the maneuver by returning to its original lane. This scenario presents a dual-layered challenge in planning. The corresponding simulation snap is presented here in Figure~\ref{fig:avoid_bike}.

In this scenario, the ego vehicle decides to overtake a cyclist ahead, considering traffic regulations and the speed of surrounding vehicles. It verifies the legality of overtaking based on dashed road markings and ensures the lane is wide enough to maintain the required 3-foot clearance. Once these conditions are met, the framework transitions from the lane-following state to the overtaking state. During the maneuver, the vehicle adheres to speed limits and maintains a safe distance. After completing the overtaking, the vehicle returns to the lane-following state. This complex maneuver demonstrates that the VLM accurately recognizes the cyclist, and the FSM effectively guides the VLM to assess the target lane conditions, showcasing the framework's coherent functionality across the VLM and FSM.  

\subsubsection{Right Turn on Red}
\begin{figure}[!ht]
  \centering
  \includegraphics[width=0.5\textwidth]{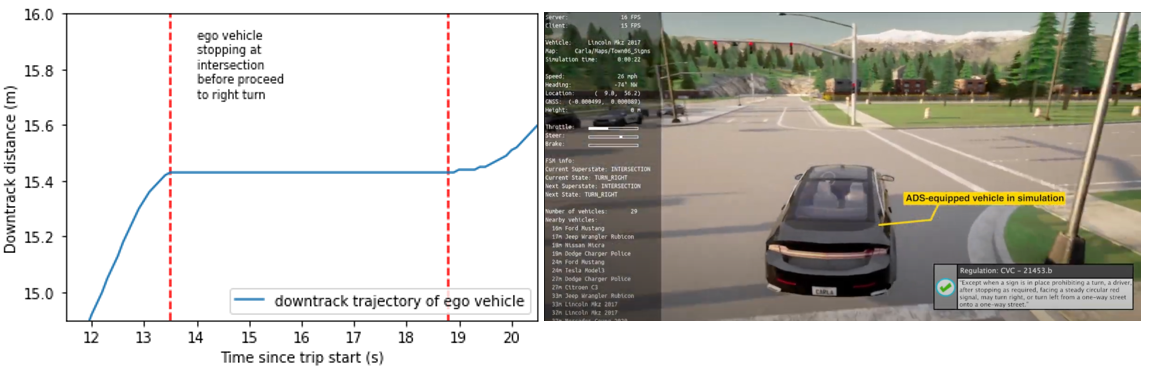}
  \caption{Trajectory plot and the simulation screenshot when ego ADS vehicle making a right turn during a red light.}\label{fig:right_red}
\end{figure}

Making a right turn on red poses a complex challenge for automated driving. The ego vehicle must detect the red light, stop at the line, and then proceed with the turn. This stop-and-check process is marked by two red dashed lines, as shown in Figure~\ref{fig:right_red}.

The ego vehicle must first confirm the legality of a right turn on red, as regulations vary by region. This requires the automated driving system to reference a comprehensive traffic rule database for the jurisdiction. Once legality is established, the vehicle ensures safety by adhering to the CVC and coming to a complete stop before turning. Using its sensors, the vehicle assesses the environment for oncoming traffic and proceeds only when safe. The process transitions from lane-following to intersection-handling, with the path planning framework checking for prohibitive signs before authorizing the turn if conditions allow. Compared to previous regulations, this scenario involves recognizing multiple traffic signs, further validating the VLM's capability in understanding complex environments, generalizing its ability to comprehend widely different objects, and demonstrating its seamless integration with the FSM.

\subsubsection{Variable Speed Limit}
In this scenario, managing variable speed limits in school zones poses a significant challenge for ADS, as they must anticipate upcoming zones. The ego vehicle, operating in the lane-following superstate, detects the school zone and begins decelerating early, marked by a red vertical line. The accurate detection of such zones—either through visual recognition or HD-map annotation—is crucial to ensure pedestrian safety, particularly for children. School zones require a controlled reduction in speed, and the vehicle must reach the correct speed limit by the time it enters the zone, ensuring smooth driving without abrupt changes. The yellow dashed vertical line shows the vehicle maintaining reduced speed through the zone, with the framework calculating proximity and speed limits accurately despite the regulation having a vagueness score of one. The system effectively adjusts the vehicle’s speed and continues the route while complying with the regulation.
\begin{figure}[!ht]
  \centering
  \includegraphics[width=0.5\textwidth]{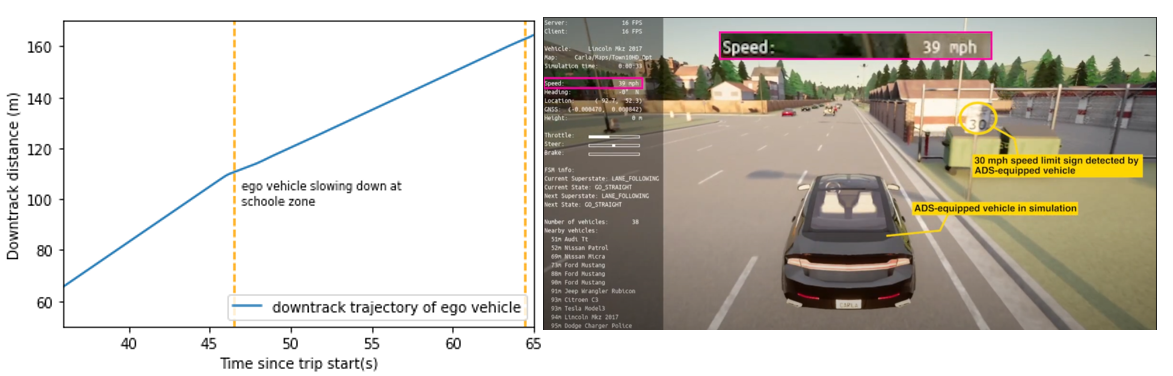}
  \caption{Trajectory plot and the simulation screenshot when ego ADS vehicle adjusting speed for areas with different speed limit regulations.}\label{fig:variable_speed_limit}
\end{figure}

\subsection{Real-World Testing}
Real-world testing adds complexity by introducing actual traffic signs and dynamic elements, including VRUs. Unlike simulations, real environments are unpredictable, requiring the VLM to accurately interpret key traffic elements in real-time. The VLM must process live data and provide precise inferences for the ADS to generate lawful and feasible driving plans. To evaluate LLaVA’s performance in identifying critical traffic scenarios, Table~\ref{tab:vlm_results} summarizes detected elements and their corresponding inference responses.

\begin{table*}[!ht]
\centering
\caption{VLM Detection and Response in Key Traffic Scenarios}
\begin{tabular}{|p{4cm}|p{4cm}|p{8.5cm}|}
\hline
\textbf{Meaningful Traffic Scenario} & \textbf{VLM Detection Result} & \textbf{VLM Response} \\ \hline
Nearby intersection & Detected & ``Ego vehicle is approaching an intersection…'' \\ \hline
Yield for pedestrian & Detected & ``The intersection appears to have pedestrian crossing, the ego vehicle should stay alert and yield to pedestrians.'' \\ \hline
Road work warning sign & Detected & ``There is a visible warning sign: Road Work Ahead.'' \\ \hline
Speed limit sign & Detected & ``The speed limit sign indicates 20 mph maximum speed.'' \\ \hline
Traffic light - red & Detected & ``There is a visible red traffic light in sight.'' \\ \hline
Traffic light - green & Detected & ``There is a visible green traffic light in sight.'' \\ \hline
Bicycle lane & Detected & ``There is a visible bicycle lane.'' \\ \hline
End road work warning sign & Miss Detected as road work sign. & ``There is a visible warning sign: Road Work Ahead.'' \\ \hline
Stop here on red sign & Miss Detected as stop sign. & ``There is a visible stop sign.'' \\ \hline
\end{tabular}
\label{tab:vlm_results}
\end{table*}

\subsubsection{VLM Inference Accuracy}
As shown in Table~\ref{tab:vlm_results}, the model successfully detected most scenarios but failed to recognize the ``end road work" and ``stop here on red" signs, correctly identifying 7 out of 9 key cases. It performed well in recognizing dynamic elements like vehicles, pedestrians, cyclists, and intersections but struggled with less common or text-heavy signs, particularly in zero-shot inference.

A key limitation of the VLM is its difficulty in distinguishing traffic signs with similar shapes and colors, such as ``road work" and ``end road work." This issue stems from the LLaVa model being a general-purpose vision-language model trained on diverse scenarios rather than a dataset specialized in traffic environments with various traffic signs. Additionally, the wide-angle camera used in testing causes traffic signs to occupy only a small portion of the image, making it harder to differentiate visually similar signs. 

\subsubsection{VLM Inference Efficiency}
The inference speed of the tested VLM is critical for real-time ADS planning. As part of the perception module, its processing rate directly influences the operational frequency of the path planning framework, with other modules running at a minimum of 5Hz. Several factors impact VLM performance, including image resolution, which affects both the level of detail and the computational load. Additionally, different models and quantization methods present trade-offs between accuracy and efficiency, requiring a balance between precision and real-time feasibility. 

\begin{table}[!ht]
\centering
\caption{LLaVA \cite{liu2024llava} inference frequency with different models and different quantization}
\begin{tabular}{|p{2.5cm}|p{2cm}|p{2.5cm}|}
\hline
\textbf{Model} & \textbf{Quantization} & \textbf{Inference Frequency} \\ \hline
LLaVA-1.5-7B & 4 bit & 2 Hz \\ \hline
LLaVA-1.5-7B & 8 bit & 0.5 ~ 1 Hz \\ \hline
LLaVA-1.5-13B & 4 bit & lower than 2 Hz \\ \hline
LLaVA-1.5-13B & 8 bit & lower than 0.5 Hz \\ \hline
\end{tabular}
\label{tab:vlm_freq}
\end{table}

Table~\ref{tab:vlm_freq} presents the inference frequencies of different LLaVA-1.5 \cite{liu2024llava} model quantizations using 1080p frames from simulation and real-world tests. The 7B model with 4-bit quantization consistently operates at 2 Hz, while the 8-bit version ranges from 0.5 to 1 Hz. The 13B model with 4-bit quantization occasionally reaches 2 Hz but generally runs slower, with its 8-bit version falling below 0.5 Hz. These results indicate that LLaVA-1.5-7B with 4-bit quantization offers the best balance between speed and efficiency. Lowering image resolution could improve inference speed but compromise critical visual details like traffic lights and signs, making planning more difficult. The VLM meets the 2 Hz requirement for strategic planning using a single NVIDIA RTX 4070 GPU. More importantly, the test confirms the feasibility of VLM integration in real-time planning, with further optimization possible through both camera and computational hardware upgrades. 

\section{CONCLUSION}
In conclusion, this study demonstrates the effectiveness of a novel, generalized ADS planning framework designed for traffic regulation compliance. Key advancements, tested in both simulated and real-world environments, include the integration of a VLM, customized prompt texts structured with the FSM, and a comprehensive regulation database. These elements work together to ensure the system accurately interprets and adheres to traffic regulations, enhancing the legality and safety of ADS planning. Overall, the proposed framework addresses existing limitations in ADS traffic law and regulation awareness. Testing results confirm the effectiveness of the proposed framework and indicate the VLM has promising potential in real-time strategic planning tasks. 

Future work will focus on fine-tuning VLMs with traffic-oriented datasets to improve scene interpretation and domain-specific accuracy, especially in cases where real-world testing reveals missed detections of visually similar traffic signs. a more specialized VLM ensures the capability of handling a more comprehensive range of traffic scenarios with greater precision. Moreover, transitioning from a zero-shot approach to a multi-shot prompting method will allow the model to better leverage prior situational knowledge and previous prompts, enhancing its ability to understand complex scenes and improving overall performance.

\section{ACKNOWLEDGMENT}
This material is based on work supported by the Federal Highway Administration Center of Excellence on New Mobility and Automated Vehicles, and National Science Foundation \# 2346267 POSE: Phase II: DriveX: An Open-Source Ecosystem for Automated Driving and Intelligent Transportation Research.

\newpage
\bibliographystyle{IEEEtran}
\bibliography{IEEEabrv}
\end{document}